\documentclass[
]{ceurart}

\usepackage{listings}
\begin{document}

\copyrightyear{2025}
\copyrightclause{Copyright for this paper by its authors.
  Use permitted under Creative Commons License Attribution 4.0
  International (CC BY 4.0).}

\conference{CLEF 2025 Working Notes, 9 -- 12 September 2025, Madrid, Spain}

\title{DS@GT at CheckThat! 2025: Evaluating Context and Tokenization Strategies for Numerical Fact Verification}

\title[mode=sub]{Notebook for the CheckThat! Lab at CLEF 2025}

\author[1]{Maximilian Heil}[%
orcid=0009-0002-6459-6459,
email=mheil7@gatech.edu,
]
\cormark[1]
\fnmark[1]

\author[1]{Aleksandar Pramov}[%
orcid=0009-0005-9049-1337,
email=apramov3@gatech.edu,
]

\fnmark[1]
\address[1]{Georgia Institute of Technology, North Ave NW, Atlanta, GA 30332}
\cortext[1]{Corresponding author.}
\fntext[1]{These authors contributed equally.}

\begin{abstract}
Numerical claims — statements involving quantities, comparisons, and temporal references — pose unique challenges for automated fact-checking systems. In this study, we evaluate modeling strategies for veracity prediction of such claims using the QuanTemp dataset and building our own evidence retrieval pipeline. We investigate three key factors: (1) the impact of more evidences with longer input context windows using ModernBERT, (2) the effect of right-to-left (R2L) tokenization, and (3) their combined influence on classification performance. Contrary to prior findings in arithmetic reasoning tasks, R2L tokenization does not boost natural language inference (NLI) of numerical tasks. A longer context window does also not enhance veracity performance either, highlighting evidence quality as the dominant bottleneck. Our best-performing system achieves competitive macro-average F1 score of 0.57 and places us among the Top-4 submissions in Task 3 of CheckThat! 2025. Our code is available at \url{https://github.com/dsgt-arc/checkthat-2025-numerical}.
\end{abstract}


\maketitle

\section{Introduction}

In the broader scope of automatic fact-verification systems, the CLEF CheckThat! 2025 edition through its different tasks, investigates best practices for each component of the typical pipeline of such a system: from establishing claim check-worthiness, undergoing claim normalization and culminating in evidence retrieval and natural language inference.\cite{clef-checkthat:2025-lncs, CheckThat:ECIR2025} Task 3 in particular, focuses on systems for automatic claim verification of numerical and temporal claims\cite{clef2025-workingnotes, clef-checkthat:2025:task3}. Numerical misinformation — claims involving statistics, comparisons, intervals, and temporal expressions — poses a distinct challenge to automated fact-checking systems. While advances in neural architectures and the availability of large-scale datasets have significantly improved claim verification for general text, verifying numerical claims remains underexplored and substantially more difficult. This is especially critical given the documented “numeric-truth effect,” where the presence of numbers lends false credibility to misinformation \cite{sagara2009numerictruth}.

To that end, CheckThat! Task 3 employs the QuanTemp dataset that was introduced as the first large-scale, real-world benchmark dedicated entirely to the verification of numerical and temporal claims \cite{venktesh2024quantemp}. Collected from 45 fact-checking organizations worldwide, it comprises over 15,000 diverse claims annotated with fine-grained labels—True, False, or Conflicting—and is accompanied by a web-sourced evidence corpus of over 423,000 specially parsed evidence snippets. This dataset spans statistical, comparative, interval, and temporal categories, offering a comprehensive resource for evaluating and improving models on numerical fact-checking tasks.
  
Our approach investigates improvements over existing methods by having two premises: 1) More evidences and longer context will improve the prediction quality of the model, provided that the model can actually handle the longer context with an increased length of the input tokens; 2) Acknowledging that the way tokenization is done on numbers plays a role in abstract reasoning for LLM models, and that a switch in the number tokenization from left-to-right to right-to-left (R2L) could aid even a BERT-based model in the veracity estimation of numerical claims \cite{lee2024digitstodecisions}. 
To this end, in our study, we postulate the following three research questions (RQs) 
\begin{description}
  \item[RQ 1] Does longer context (3 vs. 9) of retrieved evidence snippets improve the veracity prediction? 
  \item[RQ 2] Does R2L-tokenization improve performance? 
  \item[RQ 3] Does combining long context and R2L-tokenization outperform the other settings? 
\end{description}

Through ablation studies, we answer each of them by employing ModernBERT\cite{warner2024smarter} - a BERT architecture which is specifically suited to maintain longer context, as well as applying R2L numerical tokenization. In addition, the dataset exhibits an unbalanced distribution of the true/false/conflicting labels , with false claims being notably with the highest frequency\footnote{See Section~\ref{sec:eda}.}. Therefore, we also experiment with focal loss to address the mild class imbalance \cite{lin2017focal}. In addition, for fast prototyping we also employ LoRA, which adapts only a small portion of the parameters during fine-tuning \cite{hu2022lora}. 

In our modeling pipeline, we therefore conduct the in-depth analysis on QuanTemp's English dataset by building a hybrid BM25 + transformer evidence retrieval-reranking pipeline, subsequently plugged into a ModernBERT Natural Language Inference (NLI) core. Moreover, following \cite{venktesh2024quantemp}, we also employ a claim decomposition step\cite{claimdecomp2022}, which aims to split the claim into separate parts and improve the retrieval stage, in a matter akin to query expansion. 

To answer \textbf{RQ1}-\textbf{RQ3}, we run a systematic ablation that varies context length and applies R2L numerical tokenization. Our experiments—spanning ModernBERT, MathRoBERTa, and GPT-4o-mini \cite{OpenAI_GPT4_2023} few-shot prompting—show that the longer-context, R2L-tokenized ModernBERT configuration delivers the strongest performance.

The paper is structured as follows: Section~\ref{sec:rec} shows related works, Section~\ref{sec:eda} presents the explanatory data analysis (EDA), Section~\ref{sec:Modeling} explains our modeling approach, Section~\ref{sec:res} presents our results, Section~\ref{sec:future} show further research avenues and Section~\ref{sec:conclusion} concludes.
 
\section{Related work}
\label{sec:rec}
The major inspiration for our modeling pipeline (and indeed the core of the English dataset provided by CLEF) is based on the work of QuanTemp, which itself is based on an earlier version that appears in NumTemp \cite{venktesh2024numtemp}. Their contribution closes a particular gap in the field of fact-checking research - the verification of numerical claims, which are typical around e.g., political debates.  While many prior efforts have proposed automated claim verification systems and evidence retrieval architectures (e.g., \cite{hassan2017claimbuster}, \cite{chen2023complex}), these largely target general textual claims and rarely emphasize the intricacies of quantitative or temporal reasoning.

% Indeed, while the authors note that there many automatic claim verification \& relevant evidence retrieval systems (see e.g. \cite{hassan2017claimbuster},  \cite{chen2023complex}) applied on diverse real and/or synthetic datasets, there is no dataset that focuses specifically on numerical claims. At the same time, in particular around e.g., political debates, numerical claims and the need for their fact-checking increases. 

At the first step of the pipeline, \cite{venktesh2024quantemp} adopts a typical two-step evidence retrieval-reranking approach\cite{hambarde2023information}. First, it uses a sparse (BM25) retrieval system to retrieve a broader set of (100) evidences,  which then subsequently reranked by a  neural reranking model which takes into account semantic relevance on top of the key-word matching that BM25 delivers.

% AP: REWRITE
For claim verification, they experiment with several transformer-based natural language inference (NLI) models, including BERT variants, and also explore prompting strategies for large language models in both few-shot and zero-shot configurations. This dual focus on retrieval and verification for numerical claims makes their framework especially relevant for real-world applications, where numerical misinformation can be particularly persuasive due to the well-documented "numeric-truth effect." Our work builds on this foundation, extending it through targeted ablations, alternate tokenization schemes, and additional model comparisons that further explore the boundaries of model performance in this domain.
 
\section{Exploratory Data Analysis}
\label{sec:eda}

The (English) dataset consists of 432,320 evidence snippets and 15,514 claims labeled in three categories - 18.79\% \emph{True}, 57.93\% \emph{False} and 23.27\% \emph{Conflicting}. More details on the composition of the dataset and the procedure used to retrieve the evidence are outlined in QuanTemp.

It is of interest to consider the semantic overlap between the labels of Conflicting, True, and False statements. If these labels form well-separated clusters in a lower-dimensional embedding space, this could suggest that veracity prediction is more tractable—since semantically distinct groups are easier to classify. Conversely, overlapping regions between these classes may highlight ambiguous or nuanced cases that are inherently more difficult to resolve, especially for automated systems.
 
\begin{figure}[h]  
  \centering
  \includegraphics[width=0.8\textwidth, height=0.6\textwidth]{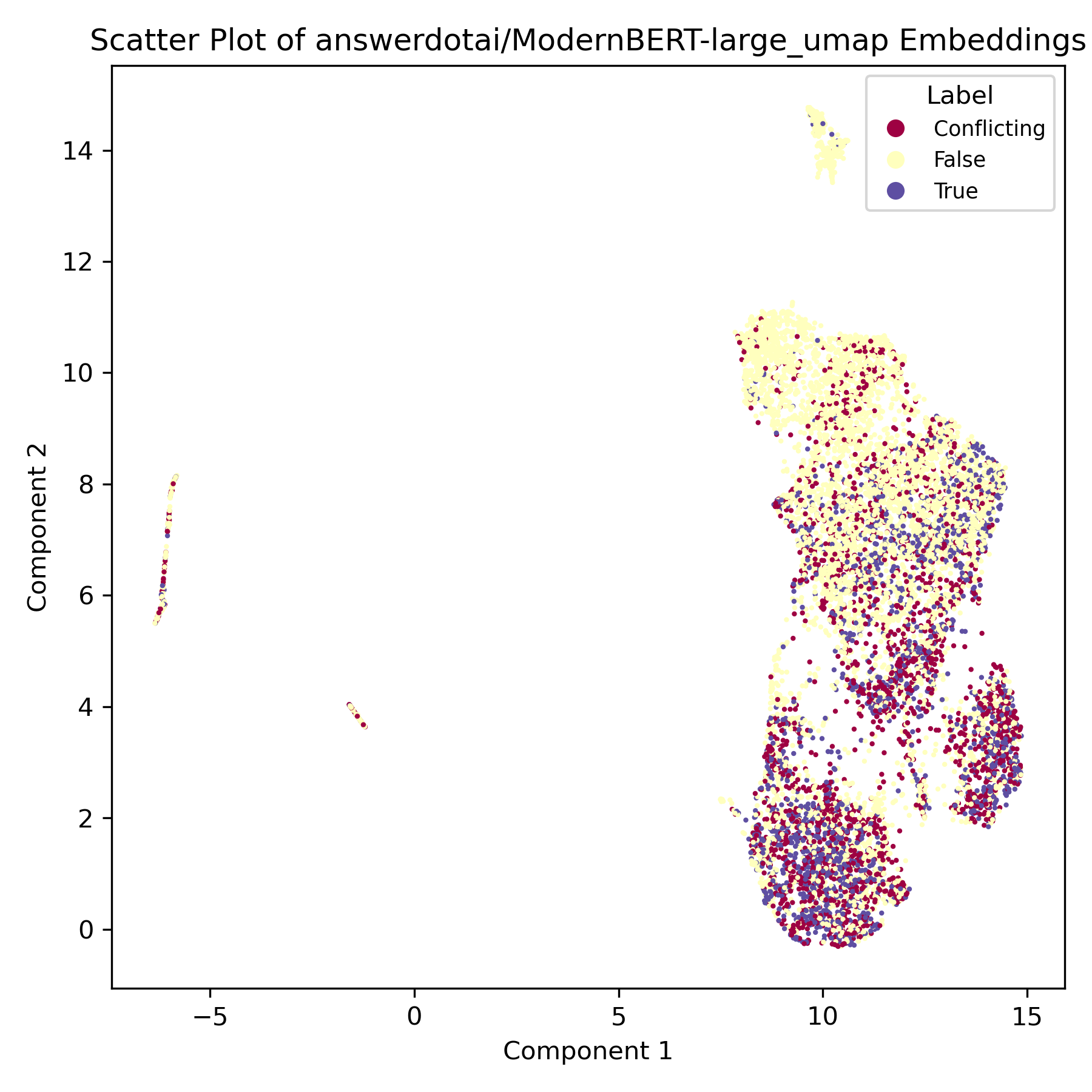} %  
  \caption{UMAP visualization of the original claims in the english train dataset, stratified by label True/False/Conflicting}
  \label{fig:umap}
\end{figure}
 
To that end, we apply a Uniform Manifold Approximation and Projection (UMAP) \cite{mcinnes2018umap} dimensionality reduction to the ModernBERT-large embeddings of the claims in the English training dataset (in their original, non-decomposed form). The resulting scatterplot visualization, shown in Figure \ref{fig:umap}, displays the embeddings colored by their respective veracity labels: True, False, and Conflicting.

The plot reveals a noticeable separation of False claims from the other two categories, while True and Conflicting statements frequently overlap. This pattern can be attributed both to the inherent difficulty of distinguishing these categories—particularly when evidence is ambiguous or contradictory—and to the complex mapping strategy employed in the QuanTemp dataset, where nuanced labels (e.g., “mostly true”) are abstracted into broader categories like Conflicting \cite[Table 9]{venktesh2024numtemp}. Such overlap in semantic space underscores the challenges in automated veracity assessment, especially for claims requiring nuanced interpretation or context-dependent evidence.
  
\section{Modeling Approach}
\label{sec:Modeling}

\begin{figure}[h]  
  \centering
  \includegraphics[width=0.8\textwidth, height=0.7\textwidth]{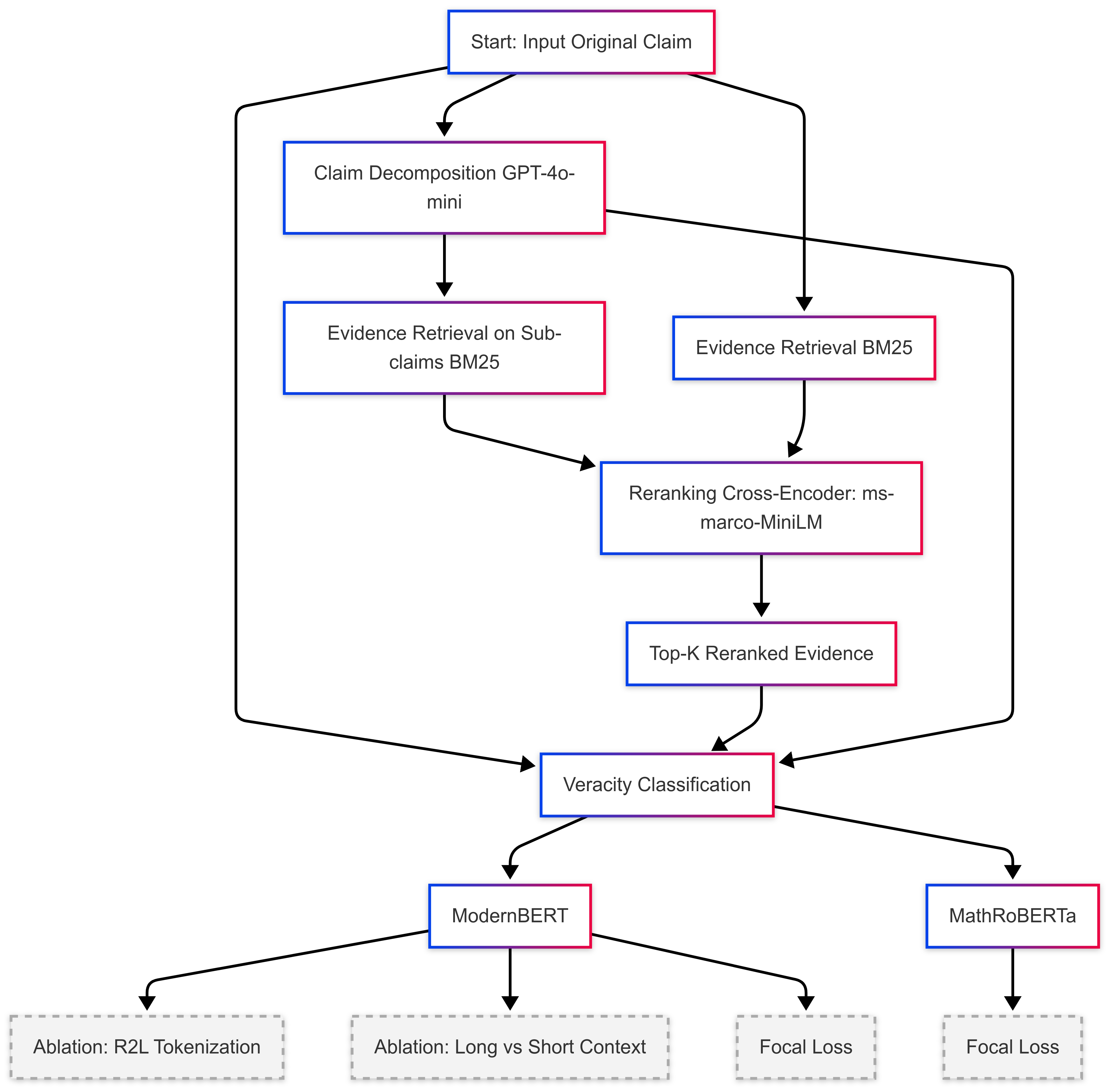} %  
  \caption{Modeling pipeline outlining the steps: (a) Claim decomposition (b) Evidence retrieval and Reranking and (c) Veracity classifier models for the natural language inference tasks}
  \label{fig:modelingpipeline}
\end{figure}

Our modeling approach follows closely the typical claim verification pipeline of evidence retrieval \& reranking, followed by a veracity classifier. Figure \ref{fig:modelingpipeline} depicts the process.

\subsection{Evidence Retrieval}
At the first step, as done in \cite{venktesh2024quantemp}, we also perform with a claim decomposition by using GPT-4o-mini \cite{OpenAI_GPT4_2023}, which aims to split the original underlying claim into 3 separate smaller claims, for which we retrieve evidence from the evidence corpus. To generate structured sub-questions from factual claims, we used a prompt tailored for decomposition. The prompt was done with low temperature (0.3) to reduce randomness and increase overall coherence in the generated yes/no questions. A moderate frequency penalty (0.6) and higher presence penalty (0.8)
encouraged semantic variety while still controlling for drift from the original claim. The maximum number of tokens (300) proved to strike a good balance between the model generating diverse but also concise sub-claims. 

Next, we followed a standard two-step approach for retrieving evidence for each decomposed sub-claim \cite{hambarde2023information}: a sparse BM25 retrieval, followed by a reranking step, for which we used cross-encoder (which encodes both the query and the retrieved document). This two-step retrieval procedure (BM25 + reranker) showed good performance in \cite{venktesh2024quantemp} and served as a motivation for our approach.

After some experimentation with transformer-based rerankers, we settled on \emph{ms-marco-MiniLM-L-12-v2} - a cross-encoder reranker, fine-tuned on MS Marco Passage ranking dataset\cite{bajaj2016ms}. The experimentation was easily facilitated via a centralized API package by \cite{clavié2024rerankers} and while we did not experiment with LLM-based reranking, this certainly remains part of our plans for the future\footnote{See Section~\ref{sec:future}.}. 
For each decomposed sub-claim, the initial BM25 retrieval was set to 50 documents, and we would then follow with the reranking step on them. The post-reranking top 1-3 of each sub-claim were then kept for the next steps, depending on the particular specification of the ablation study, in line with \textbf{RQ 1}. The input for the subsequent NLI task then constitutes an input string composed of: the original claim, (a set of the) decomposed claims and the respective retrieved \& reranked evidences.

 \subsection{Natural Language Inference}
Our main workhorse model for the veracity classifier was ModernBERT - an optimized model based on the BERT architecture, that can natively support sequence length of 8,192, a considerable increase of the usual BERT-based context window of 512. It does so by introducing changes to the embedding procedure (employing rotary positional embeddings), which are more scalable to longer contexts, as well as an alteration between global and local attention mechanisms to manage memory and compute efficiency. 
The resulting support for (much) longer context (max input length) lends itself to explore the effect of shorter vs. longer context as stated in \textbf{RQ 1} and was thus the main driver behind our modeling choice. Within the transformer based classifiers, we also experimented with MathRoberta - a fine-tuned Roberta model \cite{liu2019roberta} on mathematical discussion posts, which was part of the CheckThat! Task 3 baseline. 
Both encoders have been fine-tuned with a learning rate of 2e-5 and cross-entropy loss.
Training has been performed on two, Quadro RTX 6000-24GB, two Tesla V100-32GB or a H100-80GB on the Phoenix cluster of Georgia Tech's Partnership for an Advanced Computing Environment\cite{Pace} or locally on an Apple M3 Pro GPU-36GB and Metal Performance Shaders.

Lastly, for the dev set, we also experimented with a few-shot prompted GPT-4o-mini \cite{OpenAI_GPT4_2023} model as a classifier, on the original claims. To incorporate dataset priors into the model’s output, we apply a logit bias based on the empirical distribution of labels (True, False, Conflicting) in the validation set. Specifically, for each target label, we compute the log-odds of its prior probability and scale these values by a factor $\alpha$, which controls the strength of the prior. These scaled log-odds are then assigned as biases to the corresponding output tokens, encouraging the model to produce predictions that are nudged towards the observed label distribution in the original claims. The empirical results however, showed significantly worse performance than the BERT-based classifier and we thus did not pursue that avenue further.

% Focal loss
One additional change that we applied for the transformer-based models was \emph{focal loss} \cite{lin2017focal}.  While the imbalance in the dataset is not severe by any means, we investigated whether accounting for said imbalance could bring an added value in the performance. Focal loss addresses class imbalance by down-weighting easy examples and focusing training on hard, misclassified ones. It does so by applying a modulating factor to the standard cross-entropy loss, reducing the impact of well-classified majority class examples and thereby improving the model's sensitivity to underrepresented classes.

% R2L tokenization
As noted by \cite{lee2024digitstodecisions}, tokenization can play a critical role in numerical reasoning with large language models, as the usual left-to-right segmentation of numbers can hinder the numerical reasoning. It remains unclear whether the same effect would hold for higher-level tasks, such as NLI for numerical claims. We therefore integrate their right-to-left (R2L) tokenization into ModernBERT and evaluate whether this altered token order also improves transformer-based NLI performance on numerical fact-verification.

\subsection{Evaluation}
We fine-tune and evaluate our models based on the macro-averaged F1-measure (macro F1)
\begin{equation}
   \text{Macro-F1} = \frac{1}{N} \sum_{i=1}^{N} F1_i
\end{equation}
where $F1_i$ is the class-wise F1 score 
\begin{equation}
    F1_i = 2 \times \frac{P_i \times R_i}{P_i + R_i}
\end{equation}
where $R_i$ is the recall of class $i$ and $P_i$ is the precision of class $i$.

\section{Results}
\label{sec:res}

% \usepackage{multirow}
\begin{table}[h]
\caption{Experiment and Ablation Study Results}
\label{tab:res}
\begin{tabular}{l|cc|ccccc}
\multirow{2}{*}{\textbf{Run}} & \multicolumn{2}{c|}{\textbf{Train}} & \multicolumn{5}{c}{\textbf{Validation}} \\
 & Macro-a. F1 & Acc. & Macro-a. F1 & False F1 & Conflicting F1 & True F1 & Acc. \\ \toprule
Benchmark & 0.75 & 0.67 & 0.56 & 0.79 & 0.48 & 0.41 & 0.66 \\ \midrule
Our-Data & 0.56 & 0.67 & 0.52 & 0.80 & 0.29 & 0.46 & 0.64 \\
Short-Context & 0.50 & 0.70 & 0.52 & 0.77 & 0.42 & 0.37 & 0.61 \\
Long-Context & 0.64 & 0.74 & 0.52 & 0.78 & 0.37 & 0.41 & 0.62 \\
R2L Short-Context & 0.38 & 0.60 & 0.45 & 0.79 & 0.40 & 0.16 & 0.63 \\
R2L Long-Context & 0.42 & 0.60 & 0.47 & 0.79 & 0.32 & 0.30 & 0.62 \\ \midrule
\textbf{Submission} & \textbf{0.63} & \textbf{0.71} & \textbf{0.57} & \textbf{0.81} & \textbf{0.36} & \textbf{0.55} & \textbf{0.66} \\ 
PEFT & 0.49 & 0.63 & 0.49 & 0.78 & 0.30 & 0.37 & 0.63 \\
Focal-Loss & 0.65 & 0.75 & 0.57 & 0.81 & 0.41 & 0.50 & 0.67 \\ \bottomrule
\end{tabular}
\end{table}

%Baseline
Following our research questions, Table \ref{tab:res} presents our results. First, we show results obtained with the QuanTemp dataset and natural language inference with MathRoBERTa (Benchmark). The macro-avg. F1 drops from train (0.75) to validation (0.56) indicating weak generalization and an overfit after 3 epochs of fine-tuning. Train accuracy (0.67) and validation accuracy (0.66) remains stable as it is less sensitive to the class distribution.
%Our recreation of venktesh
When we recreate the claim decomposition and the evidence retrieval but stick with MathRoBERTa (Our-Data), we observe a slight performance drop for train (macro-avg. F1: 0.56) and validation (macro-avg. F1: 0.52). This is especially driven by less and fewer correct classification of conflicting claims. Apparently, our recreation of the numerical claim verification pipeline provides poorer evidences for the encoder to verify the claims during natural language inference.
% Context window

Next, we explore \textbf{RQ 1} by comparing short (256 tokens) and long (1,024) context windows given our dataset with up to 3 evidences per question and ModernBERT, an encoder with a maximum context window of 8,192 tokens. Here we contrast 1 evidence per question and 256 context window (Short-Context) with 3 evidences per question and a 1,024 context window (Long-Context). Whilst training performance between short-context (macro-avg. F1: 0.50) and long-context (macro-avg. F1: 0.64) vary significantly, validation performance is similar. This indicates that a longer context window is not helpful for numerical veracity prediction. Hence, our results agree with findings in \cite{venktesh2024quantemp}. Nevertheless, our results could be limited by poorer evidences for the veracity prediction. Ultimately, providing the inference system with three weak evidences instead of one weak evidence could prove to be without an effect on the final outcome.
% R2L tokenization

Moreover, we investigate \textbf{RQ2} the effect of tokenization by switching the ModernBERT Tokenizer to Right-to-left (R2L) instead of Left-to-right. Results are based on our retrieved evidence dataset. R2L Short-Context achieves a 0.38 macro-avg. F1 for train and a 0.45 macro-avg. F1 for validation. Similar, R2L Long-Context also performs poor and macro-avg. F1 for train (validation) drops to 0.42 (0.47). These results contradict findings in \cite{lee2024digitstodecisions}. This surprise in performance can be attributed to the vast difference of use cases between arithmetic tests of language models compared to our numerical and temporal veracity prediction in CheckThat! Task 3. Here, the numerical aptitude of the language model might not be so critical compared to tests of arithmetic. In addition, these findings support the conclusion that the quality of the retrieved evidences is poor; high-quality evidences are vital for high-performance veracity prediction. 

These results also show a negative outcome for \textbf{RQ3}, the combination of long-context and R2L tokenization show underwhelming performance.
% Submission
These results motivate us to fine-tune ModernBERT on the QuanTemp dataset (Submission) for submission. Here, we do not employ R2L tokenization and just use 1 evidence snippet per question. Nevertheless, we keep the context-window at 1,024 tokens to empower the encoder to attend to all tokens during the veracity prediction. This results in a validation performance similar to the organizer's benchmark (macro-avg. F1: 0.57). Whereas our submission-model has better performance when classifying true claims (validation F1: 0.55) and weaker performance when classifying conflicting claims (validation F1: 0.36).
% Additional: PEFT & Focal
We extend our research by exploring parameter-efficient fine-tuning (PEFT) and focal loss. As expected, ModernBERT with PEFT on the QuanTemp dataset has a weaker performance as validation macro-avg. F1 drops to 0.49. 
Results with focal loss are not statistically significant from the submission results with cross-entropy loss. ModernBERT trained on the organizer dataset with focal loss achieves 0.65 macro-avg. F1 for train and 0.57 macro-avg. F1 for validation. This demonstrates that focal loss is not helpful in case of weak class imbalance. \\

\begin{table}[h]
\caption{Fine-tune Time Complexity}
\label{tab:time}
\begin{tabular}{l|ccc}
\textbf{Run} & \textbf{Runtime (min.)} & \textbf{Epochs} & \textbf{Time Efficiency} \\ \toprule
Benchmark & 6.97 & 5 & 1.39 \\ \midrule
Our-Data & 9.45 & 7 & 1.35 \\
Short-Context & 12.70 & 6 & 2.12 \\
Long-Context & 30.83 & 5 & 6.17 \\
R2L Short-Context & 5.62 & 2 & 2.81 \\
R2L Long-Context & 15.65 & 2 & 7.83 \\ \midrule
\textbf{Submission} & \textbf{13.07} & \textbf{3} & \textbf{4.36} \\
PEFT & 17.68 & 4 & 4.42 \\
Focal & 28.22 & 4 & 7.06 \\ \bottomrule
\end{tabular}
\end{table}

Table \ref{tab:time} shows the time complexity of fine-tuning the different scenarios with a NVIDIA H100-80GB. We report runtime in minutes, epochs until fine-tuning is finished (early-stopping after 2 epochs) and finally the ratio out of runtime and epochs (time efficiency). We see that the smaller model, MathRoBERTa, typically finishes fine-tuning earlier (1.35 - 1.39 ratios) than ModernBERT approaches (2.12 - 7.83 ratios). Prolonged Fine-tuning with ModernBERT is driven by the quadratic costs of attention when using a large context window. Surprisingly, when only fine-tuning 1\% of the trainable parameters of the encoder via PEFT, the time efficiency does not drop. This shows that our train pipeline has significant overhead which eliminates all the time complexity benefits of PEFT.
\\

\begin{table}[h]
\caption{Test Results with Submission Run}
\label{tab:test}
\begin{tabular}{lcccc}
\textbf{} & \textbf{Macro-Avg. F1} & \textbf{False F1} & \textbf{Conflicting F1} &  \textbf{True F1} \\ \toprule
Submission & 0.52 & 0.80 & 0.39 & 0.38 \\ \bottomrule
\end{tabular}
\end{table}
 
Finally, Table \ref{tab:test} presents our test results placing us on the $4^{th}$ out of 10 participants. It documents a further drop in performance from validation (macro-avg. F1: 0.63) to test (macro-avg. F1: 0.58). This is driven by poorer precision and recall of the true claim class (validation F1: 0.36, test F1: 0.38).
\section{Future work}
\label{sec:future}
Our findings point to several directions for future improvement. Replacing or augmenting BM25 with hybrid retrieval methods — combining dense and sparse representations - and incorporating LLM-based reranking could enhance evidence quality, which remains a key bottleneck in our pipeline. Both, reranking models and the ModernBERT classifier, could benefit from additional fine-tuning on math- and number-centric corpora (e.g., MathQA~\cite{mathqa}) to strengthen numerical reasoning. Furthermore, normalizing numbers, dates, and other numerical or temporal expressions in both claims and evidence may help language models better capture semantic equivalence and distinctions in embedding space, thereby improving inference precision. Lastly, leveraging ensembles of veracity classifiers could improve robustness by integrating complementary strengths across model architectures and training configurations.

\section{Conclusion}
\label{sec:conclusion}

Our paper presents our modeling choices for numerical fact verification in Task 3 of the CheckThat! 2025 Lab. We evaluated the impact of longer input context windows, right-to-left (R2L) numerical tokenization, and their combination on the veracity prediction of numerical and temporal claims. After recreating the evidence retrieval pipeline with claim decomposition, we observed a drop in performance suggesting a weaker quality of evidences. Therefore, in absence of high-quality evidence, we show that neither longer context nor R2L tokenization improves performance. Contrary to our expectations, this suggest that extending input size or altering tokenization strategy is less important than constructing a high-quality evidence retrieval pipeline.

Our strongest results were achieved by a ModernBERT-based pipeline using only one evidence snippet per decomposed question, a 1,024 context length and classical left-to-right tokenization. This configuration achieved competitive validation performance and placed us among the top-performing systems in the shared task. However, the overall drop in performance from validation to test highlights generalization challenges in this domain.

Our results motivate future efforts in hybrid retrieval, numerical normalization, and ensemble modeling to further innovate in the underexplored subfield of numerical claim verification.

\begin{acknowledgments}
We thank the DS@GT CLEF team for providing valuable comments and suggestions.
This research was supported in part through research cyberinfrastructure resources and services provided by the Partnership for an Advanced Computing Environment (PACE) at the Georgia Institute of Technology, Atlanta, Georgia, USA.
\end{acknowledgments}

\section*{Declaration on Generative AI}
During the preparation of this work, the authors used OpenAI-GPT-4o: Grammar and spelling check. After using this tool, the authors reviewed and edited the content as needed and take full responsibility for the publication’s content.

%%
%% Define the bibliography file to be used
\bibliography{sample-ceur}

\appendix

\end{document}